\documentclass[a4paper,10pt,twoside]{article}

\usepackage{clin}        
\usepackage{harvard}     

\usepackage{url,mdwlist,booktabs,color}
\usepackage[pdftex,pdfborder={0 0 0},linkcolor=red,urlcolor=blue,colorlinks,unicode,breaklinks,hyperfootnotes=false]{hyperref}

\begin{document}

\title{LAF-Fabric: a data analysis tool for Linguistic Annotation Framework with an application to the Hebrew Bible}

\author{Dirk Roorda$^*$ $^{**}$ \email{dirk.roorda@dans.knaw.nl}\\
{\normalsize \bf Gino Kalkman}$^{***}$ \email{g.j.kalkman@vu.nl}\\
{\normalsize \bf Martijn Naaijer}$^{***}$ \email{m.naaijer@vu.nl}\\
{\normalsize \bf Andreas van Cranenburgh}$^{****}$ $^{*****}$ \email{andreas.van.cranenburgh@huygens.knaw.nl}
\AND \addr{$^*$Data Archiving and Networked Services - Royal Netherlands Academy of Arts and Sciences, Anna van Saksenlaan 10; 2593 HT Den Haag, Netherlands}
\AND \addr{$^{**}$The Language Archive - Max Planck Institute for Psycholinguistics, Wundtlaan 1; 6525 XD Nijmegen, Netherlands}
\AND \addr{$^{***}$Eep Talstra Centre for Bible and Computing - VU University, Faculteit der Godgeleerdheid; De Boelelaan 1105; 1081 HV Amsterdam, Netherlands}
\AND \addr{$^{****}$Huygens Institute for the History of the Netherlands - Royal Netherlands Academy of Arts and Sciences, P.O. box 90754; 2509 LT; Den Haag, Netherlands}
\AND \addr{$^{*****}$Institute for Logic Language and Computation - University of Amsterdam, FNWI ILLC Universiteit van Amsterdam; P.O. Box 94242; 1090 GE Amsterdam, Netherlands}
}

\maketitle\thispagestyle{empty} 


\begin{abstract}
The Linguistic Annotation Framework (LAF) provides a general, extensible stand-off markup system for corpora.
This paper discusses LAF-Fabric, a new tool to analyse LAF resources in general
with an extension to process the Hebrew Bible in particular.
We first walk through the history of the Hebrew Bible as text database in decennium-wide steps.
Then we describe how LAF-Fabric may serve as an analysis tool for this corpus.
Finally, we describe three analytic projects/workflows that benefit from the new LAF representation:

1) the study of linguistic variation: extract cooccurrence data of common nouns between the books of the Bible (Martijn Naaijer);
2) the study of the grammar of Hebrew poetry in the Psalms: extract clause typology (Gino Kalkman);
3) construction of a parser of classical Hebrew by Data Oriented Parsing: generate tree structures from the database (Andreas van Cranenburgh).
\end{abstract}

\section{The Hebrew Bible}
The Hebrew Bible is written in old forms of Hebrew and Aramaic, which are dead languages now.
Written manuscripts are our only source for studying these languages.
The study of such a body of historical texts involves research questions from different disciplines.
Linguistic analysis is a stepping stone which must be followed by questions at higher levels of abstraction,
such as literary questions: how did authors use the system of the language to craft their design:
i.e. style, literary effect, focus, and all those features of the text that are not
dictated by the language system \cite{Peursen2010}?
Another line of questions falls into historical linguistics:
systematically charting linguistic variation in the biblical linguistic corpus can help addressing the question
as to whether the variation reflects diachronic development \cite{Peursen2006}.

\subsection{Bible and Computer}
Naturally, a research program as mentioned above seeks to employ digital tools.
In fact, a group of researchers in Amsterdam started compiling a text database in the 1970s:
the Werkgroep Informatica Vrije Universiteit (WIVU).
This resulted in a database of text and markup, the so-called WIVU database \cite{Talstra2000},
which became widely known because it became incorporated in Bible study software packages.
The WIVU markup is based on observable characteristics of the texts and refrains from
being committed to a particular linguistic theory or framework.
There is no explicit {\em grammar} to which the marked-up text material has to conform.
One of the consequences is that the data cannot conveniently be described in one hierarchical structure,
although hierarchy plays a role.
There are several, incompatible hierarchies implicit in the data.

In the 1990s the ground work has been laid for an analytical tool operating on the WIVU data.
In his PhD. thesis, \citeasnoun{Doedens1994} defined a database model and a pair of query languages
to deal with {\em text databases}, i.e. databases
in which the essential structural characteristics of text, sequence and embedding, are catered for in a natural way.
One of the query languages, QL\footnote{QL just means {\em Query Language}.}, he characterised as {\em topographic}
meaning that a query instruction is isomorphic to its results with respect to sequence and embedding \cite[pp.~108]{Doedens1994}.\footnote{The
other query language, LL, is a {\em logical} one, introduced for theoretical purposes \cite[pp.~199]{Doedens1994}.}
QL has been turned into an efficient implementation
by \citename{Petersen2004}~\citeyear{Petersen2004,Petersen2006}, allbeit in a somewhat restricted form, named MQL (Mini-QL).
The resulting database system, EMDROS \cite{Petersen2002-2014}, has become part of the WIVU tool set
and is still maintained by its author.

Between the initial work of the WIVU group then and the activities of the ETCBC group
now\footnote{With the retirement of founding father Eep Talstra in 2013, WIVU has been rebaptized as
{\em Eep Talstra Centre for Bible and Computing}.},
the web has arrived and the practice of information sharing and standardizing has changed dramatically. 
Now, in the 2010s, the need is increasingly felt to preserve and share the results of the ETCBC with
not only theologians but also with linguists and researchers in the digital humanities.
Conversely, the methods developed in the sphere of computational linguistics hold promises for new kinds of studies
of the Hebrew texts.
A first step has been the deposition of the database in its EMDROS form at DANS \cite{Talstra2012}.

\subsection{Text, markup, and annotation}
In 2013 the CLARIN-NL\footnote{\url{http://www.clarin.nl}} project SHEBANQ\footnote{System for HEBrew text:
ANnotations for Queries and Markup,
\url{http://www.godgeleerdheid.vu.nl/nl/nieuws-en-agenda/nieuwsarchief/2013/jan-mrt/clarin-subsidie-voor-hebreeuwse-database-op-internet.asp}}
started to convert the Hebrew Text Database from its EMDROS representation
into Linguistic Annotation Framework (LAF) \cite{Ide2012}.

Linguistic Annotation Framework defines a fairly generic data model for annotated data.
The assumption is that there is an immutable data resource with identifiable regions in it.
Regions can be attached to nodes, nodes can be linked with edges, and both nodes and edges
can be the targets of annotations,
which contain feature structures. This annotated graph, anchored to primary data, can be expressed in XML, by means
of the GrAF\footnote{Graph Annotation Format, \url{http://www.xces.org/ns/GrAF/1.0/}} schemas,
suggested by the LAF standard.

\begin{figure}[htb]
\includegraphics[scale=0.3]{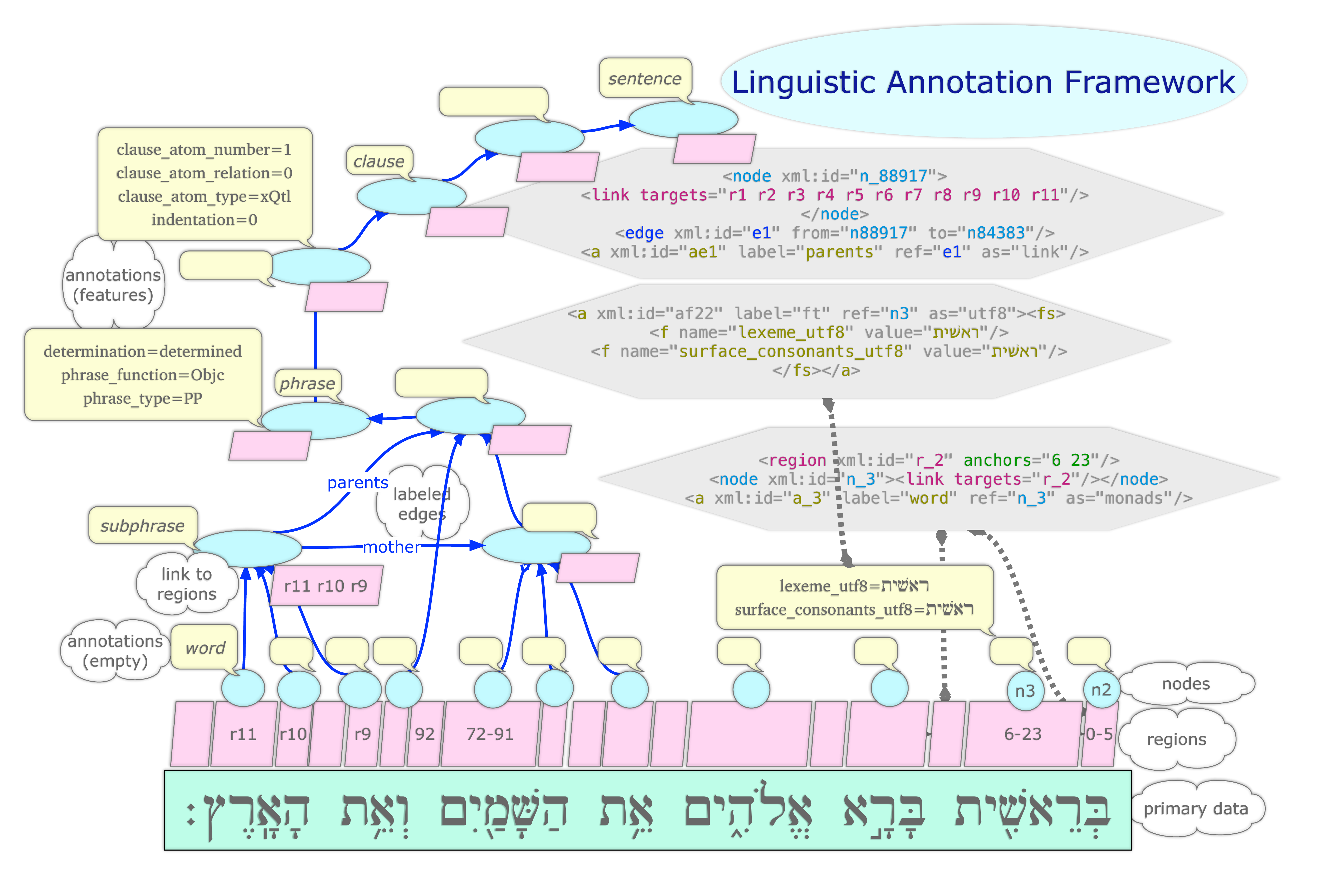}
\caption{The Hebrew text database in LAF}
\label{wivu2laf}
\end{figure}

This model has turned out to be a very good fit for the EMDROS data, which consists of objects carrying features,
where features define properties or point to other objects.
In EMDROS, objects are linked to arbitrary subsets of words,
and the concatenation of all words is the text of the Hebrew Bible
according to the Biblia Hebraica Stuttgartensia Edition \cite{BHS},
which can be regarded as an immutable resource in practice. 
Figure~\ref{wivu2laf} shows the mapping in a nutshell.

Note that this kind of markup is {\em stand-off} markup, as opposed to {\em inline markup},
of which TEI\footnote{Text Encoding Initiative, \url{http://www.tei-c.org/index.xml}} is the prime example.

\section{LAF-Fabric}
The primary result of the conversion of the Hebrew Text Database to LAF \cite{Peursen2014}
is the fact that now a standard representation of the
data can be archived. Moreover, we can preserve queries on the database as well,
following an idea expressed in \cite{Roorda2012}.
But as soon as one turns to the LAF representation, the question comes to mind: are there tools with which
this LAF resource can be processed?

\subsection{LAF tools}
There are emerging tools to deal with stand-off annotations and markup.
But here the versatility of stand-off markup must be paid for:
there are no tools that are applicable to all stand-off resources.
Even if we restrict to LAF, there are no mature tools that deal with LAF in full generality.
There are candidates, though. Here are some options and experiences.

\begin{enumerate}
\item The eXist database engine\footnote{\url{http://www.exist-db.org}}.
Initial experiments showed clearly that eXist is not designed to handle large LAF resources well
in its default configuration. Apart from an initial load time of more than an hour, even simple queries
took dozens of minutes. Surely this could have been improved by setting up indexes, but it is by no means obvious
which indexes are needed for all possible queries. We decided not to pursue this path.
\item ELAN\footnote{Eudico Linguistic ANnotator is an annotation tool that allows the
creation, editing, visualizing and searching of annotations for video and audio data.
Software developed at Max Planck Institute for Psycholinguistics, Nijmegen.
\url{http://tla.mpi.nl/tools/tla-tools/elan/}.} is a tool for annotating audio and video material.
In theory it could also be used for annotating plain text streams, but it is not designed for that.
There are modelling issues, and probably performance issues as well.
Discussions with the ELAN people at the The Language Archive\footnote{Max Planck Institute for Psycholinguistics, Nijmegen,
\url{https://tla.mpi.nl}}
made clear that ELAN works well with big primary data (audio and video) and sparse annotation data.
Our case is the exact opposite: small primary data (plain text) and very rich annotation data.
\item Graf-python \cite{Bouda2013} (component of POIO \cite{Bouda2012})
is a Python library designed to analyse generic LAF resources,
such as the
Open American National Corpus\footnote{The OANC is the corpus that drove much of the specifications of LAF.
See \url{http://www.americannationalcorpus.org/OANC/index.html}}.
However, in its current form it does not scale up to match the size of the Hebrew Text Database.
The load time is half an hour, the memory footprint is 20 GB, and these costs are incurred every time before you run
even the simplest analysis script. However, graf-python is appealing, and the first author decided to 
implement it in a new way with performance in mind.
Whereas graf-python is a clean translation of the LAF concepts into object-oriented Python code,
an efficient processor would need to compile LAF concepts into efficient data structures,
such as Python {\em arrays}, which have C-like performance.
\end{enumerate}

\subsection{Computing with LAF-Fabric}
A new, graf-python-like library was needed, one that could handle the full size of the Hebrew text database
($> 400$K words, $> 1.4$M nodes, $> 1.5$M edges, $> 33$M features) with ease.
This is not, by today's standards, a large corpus, but as it represents a fixed cultural artefact, 
it is all the data we have got and we want to examine all of them very closely.
That is why have invested in a system that allows us to manipulate the full data in main memory:
LAF-Fabric \cite{Roorda2013-2014},
a Python package that compiles LAF into a compact, binary form that loads fast.
It has an API that supports walking over the node set along edges or in various orders,
absorbing the features of the nodes in passing.

This kind of processing could also be done with a graph database, such as NEO4J\footnote{\url{http://www.neo4j.org}}.
The advantages of LAF-Fabric are the ease of installation, compactness of the compiled data, and the integration
in the Python world of scientific computing, e.g. the
{\em IPython notebook}\footnote{A quick introduction of IPython can be found at \url{http://ipython.org/ipython-doc/stable/notebook/notebook.html}} \cite{Perez2007}.

The baseline mode of data access is to walk over node sets and use edges to explore the neighbourhood of nodes.
Each node is loaded with features, from which their text and linguistic properties may be read off.
From this graph one can populate tables, vectors, trees and graphs with the data one wants to focus on.
Then, leaving LAF-Fabric, but still in the IPython notebook,
one can use the facilities of the Python ecosystem to do data analysis and visualization.
All these steps are shown in a tutorial notebook that charts the frequency of masculine and feminine gender 
for all chapters in the Hebrew Bible\footnote{\url{http://nbviewer.ipython.org/github/ETCBC/laf-fabric/blob/master/examples/gender.ipynb}}.

LAF-Fabric does not introduce a new query language, so the user of LAF-Fabric does not have to learn the ins and outs of
a new formalism.
Instead, it offers programmatic access to the nodes, edges and features in a LAF resource, in such a way that
the programmer is not burdened with the technicalities of the LAF data representation in XML.
But LAF-Fabric is not an end user tool.
End user tools are usually built on the basis of a relatively fixed concept of what results users typically want to achieve,
and those results are then offered by the tool in a user-friendly manner.
As soon as the user's needs move outside the scope of the tool in question, the user-friendliness is over.
In contrast, the approach of LAF-Fabric is that it just makes the data available,
leaving it to the user how to extract that data and what to do with it.

Clearly, LAF-Fabric is not a tool for the non-programming end user, it is best used by a {\em team} in a laboratory context,
where some members define the research needs and another member 
translates those needs into working IPython notebooks.
 
\subsection{Beyond pure LAF}
LAF is a rather loose standard. There are many ways to model specific language resources,
and it is not obvious which choices will work out best for data processing.
In particular, the semantics of edges in LAF is completely open.

Specific resources may have more structure than can be exploited by a generic LAF tool.
That is why LAF-Fabric has hooks for third-party modules to exploit additional regularities.
We ourselves developed a package called {\em etcbc} that provides extra functionality for the Hebrew Text Database
in its LAF representation:

\begin{enumerate}
\item Node ordering: the Hebrew Bible data has object types by which nodes can be ordered
more extensively than on the basis of the generic LAF data only.
LAF-Fabric can be instructed to use the {\em etcbc} sorting instead of its default sorting.
\item Data entry: {\em etcbc} contains facilities to generate forms for data entry,
read them back and convert the results to proper LAF files, which can then be added to the original resource.
\item MQL queries: {\em etcbc} has a facility to run queries on the EMDROS version of the data, and collect the results as node sets in LAF.
This gives the best of two worlds: topographic queries intermixed with node walking.
\end{enumerate}

The github repository {\em laf-fabric-nbs}\footnote{\url{https://github.com/ETCBC/laf-fabric-nbs}}
contains quite a few examples of Hebrew-LAF data processing in various degrees of maturity.

\subsection{Notes on the choice for LAF}
When the SHEBANQ project was submitted, CLARIN required that a standard format be selected from a
list\footnote{\url{http://www.clarin.eu/sites/default/files/Standards\%20for\%20LRT-v6.pdf}}.
LAF was the only obvious choice from that list.
But has it been a good choice? FoLiA \cite{Gompel2013}, which is not on the CLARIN list, comes to mind.

So far, LAF is serving us well.
Because the data model is very general, we can easily translate our data into LAF,
even where they are organized in multiple hierarchies or no hierarchy at all.
There are also features that do not represent linguistic information, but e.g. orthographic
information\footnote{The Hebrew script poses complexities: the basic information is in the consonants,
the vowels have been added later as diacritics, and there are also prosodic diacritics present.
The database provides a number of different representations for each word: with and without diacritics,
in UNICODE Hebrew or in Latin transcription.}
and numbering information, some of which are clearly ad-hoc.

All this data can be represented easily in LAF without introducing tricks and devices to work around the
specifics of the LAF data model.
We are confident that we will be able to represent the future data output of biblical scholars as well.

It must be said, however, that because of the very abstractness of LAF,
it was not obvious at first how to choose between the numerous ways in which one can represent annotation data in LAF.

FoLiA has the characteristics of an attractive ecosystem of data, formats and tools for linguistic analysis of corpora.
There are, however, a few things to be wary about when making a choice:
\begin{enumerate}
\item the ETCBC data is not the product of main-stream linguists, and there are other concerns than linguistic ones;
\item biblical scholars are continuously producing new data, in the form of new annotations.
Within the stand-off paradigm it is easy to incorporate these data in a controlled way, indicating the provenance. Any format that relies on inline markup is at a
disadvantage here;
\item the ETCBC data is essentially one document, and must fit in memory, together with a large subset of its annotations.
It seems that FoLiA is geared to corpora with multiple, smaller documents.
\end{enumerate}

\noindent Nevertheless, it is an interesting exercise to convert the LAF version of the Hebrew Bible into FoLiA,
but we leave it to one of our readers.

Graf-python has an advantage over LAF-Fabric: it can deal with feature structures in full generality.
LAF-Fabric only deals with feature structures that are sets of key-value pairs.

\subsection{Preprocessing for research}
The next sections describe several lines of research that benefit from the incarnation of the data in LAF and from
LAF-Fabric as a preprocessing tool.
They are examples of historical, literary and linguistic research lines
and in that way they serve to indicate the breadth of the landscape of biblical scholarship.
Rather than pursuing those lines in full depth,  
the purpose of this paper is to convey the importance of having good preprocessing tools, based on a standard format.
If that is in place, research efforts and results get a boost.

\section{Linguistic variation: extracting cooccurrences}
For a long time scholars have held the consensus that most of the linguistic variation in Biblical Hebrew
can be explained by assuming that there is an early variety
(Early Biblical Hebrew or EBH, in use before the Babylonian exile in the 6th century BCE)
and a late variety (Late Biblical Hebrew, or LBH,
in use after the exile)\footnote{ An early voice propagating this view is \citeasnoun{Gesenius};
nowadays one of the most influential authors using the diachronic model is Avi Hurvitz,
for instance \cite[pp.~17-34]{Hurvitz}.}.
EBH can be found mainly in the books of the Pentateuch and the Former Prophets,
whereas LBH can be found mainly in the undisputed late books
(Esther, Daniel, Ezra, Nehemiah and Chronicles).
To this diachronic model some have added dialectical variation \cite{RendsburgLE} and other kinds of
variation \cite{RendsburgDAH}.
However, in the past two decades serious challenges have been brought forward by several scholars.
Biblical Hebrew seems to be pretty homogeneous in general and several of the methods used by scholars
to study linguistic variation in Biblical Hebrew have become questionable
now it has been shown that many linguistic features of which it was thought
earlier that they are characteristic of LBH occur throughout the Hebrew Bible.
An important methodological problem is the linguistic-literary circularity involved
in many studies applying the diachronic model: a feature occurs mainly in late texts,
therefore the feature is late and this proves that the texts in which it occurs are late and
so on \cite[vol.~1,~ch.~3,4]{YoungEtAl}.
These issues have led some to propose a new model, namely that the variation in Biblical Hebrew
fits a situation in which two styles were used, a more conservative one (formerly called EBH) and a
freer one (formerly called LBH).
According to this model, both styles were used before and after the exile \cite[vol.~2,~ch.~2]{YoungEtAl}.

In the NWO-funded project {\em Does Syntactic Variation reflect Language Change?
Tracing Syntactic Diversity in Biblical Hebrew Texts}\footnote{See
\url{http://www.nwo.nl/en/research-and-results/research-projects/19/2300177219.html}},
another approach is advocated.
Instead of starting with assumptions about where or when a specific biblical text was written,
first the distribution of a large quantity of syntactic features and the way they vary throughout
the Hebrew Bible will be mapped and only after this is done the question arises to what historical,
geographical or cultural factors the resulted variation may be related.

\begin{figure}[htb]
	\begin{center}
	\includegraphics[scale=0.75]{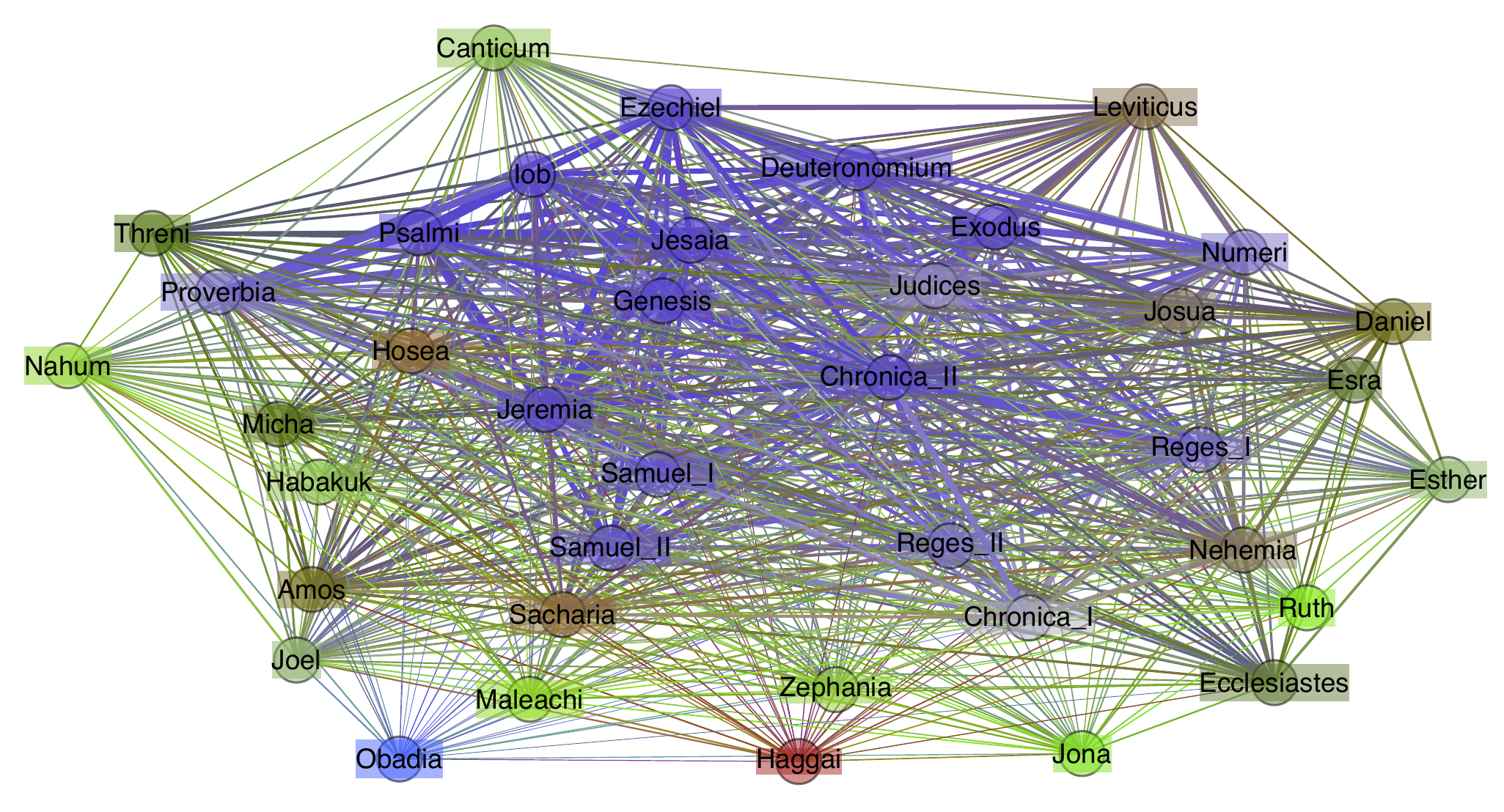}
	\end{center}
	\caption{Gephi force atlas of distribution of common nouns and verbs in the books of the Hebrew Bible}
	\label{commonnouns}
\end{figure}

Using LAF-Fabric, the first and third authors did a pilot to get a general impression of the
linguistic variation in Biblical Hebrew.
First a list was made of the lexemes of all the verbs and common nouns in the Hebrew Bible
and with this list a table was made in which the presence or absence of these lexemes
in the separate biblical books is registered.
With the data in this table a graph was made using the Force Atlas algorithm\footnote{This
is a so called force-directed algorithm, which assigns attractive and repulsive forces to the edges.
The algorithm performs a simulation of a physical system, resulting in intuitively understandable graphs.}
implemented in Gephi.
The result is shown in figure~\ref{commonnouns}.
The figure shows several expected linguistic results, such as the relatively close relationship
between the LBH books\footnote{ This approach is derived from the author's MA thesis \cite{Naaijer2012}.}. 

Although this is only a preliminary result, with the help of LAF-Fabric in combination with high level tools
for statistical analysis and visualization like Gephi or Matplotlib
(an implementation of various visualization algorithms in Python) it is possible to
analyse the use and distribution of large quantities of data, instead of focusing exclusively on details
in separate features, as is usual in biblical studies.
It may be expected that in the coming years such an approach will lead to many new results,
insights and research ideas in the study of Biblical Hebrew.

\section{Grammar of Hebrew Poetry: extracting clause typology}
For centuries Hebraists have been studying the verbal forms used in Biblical Hebrew.
Though many have tried to provide a coherent description of verbal functions in Hebrew, consensus has never been reached.
This is especially true for the functioning of the verbal forms in the poetic parts of the Hebrew Bible,
which has repeatedly been characterized as completely irregular.
Illustrative in this regard is the comment made by the grammarian \citeasnoun[pp.~29-35]{Bergstrasser},
who speaks of a {\em v\"olligen Verwischung der Bedeutungsunterschiede der Tempora}
('{\em a complete blurring of functional distinctions between the tenses}')
in Hebrew poetry
and identifies the poetic use of verbal forms as {\em regellos} ('{\em random}') and {\em ohne ersichtlichen Grund}
('{\em without apparent motivation}'). 

In a PhD-research project by the second author started in 2011, this rather {\em desperate} view on the verbal system
of Hebrew poetry
is considered unacceptable and the search for a linguistic system
regulating the poetic use of verbal forms has been taken up again.
A central assumption in the project is that the meaning of verbal forms in Hebrew is not to be described
in terms of the traditional categories of tense, aspect and mood,
but has more to do with the structuring of discourse.
Therefore, one should not focus on the bare verbal forms,
but rather on the clauses in which they are embedded and the position of these clauses in the whole of the text.
In this type of approach, which is usually defined as {\em text-linguistic},
special attention is paid to the patterns constituted by subsequent verbal forms (and their clauses).
Our research has shown that the connection of mother and daughter clause,
in particular, has a strong influence on the exact functions adopted by the verbal forms in Hebrew.
Though the forms can be assigned certain default functionalities,
the exact concretization of these basic functions can only be determined on the basis of
a detailed analysis of the broader clause patterns in which a specific clause takes its position.

In this research project we reject the tendency among Hebraists to assume a gap
between the use of verbal forms in prose and poetry.
Instead of identifying different verbal systems for the two genres or even characterizing
the use of verbs in Hebrew poetry as being devoid of any system,
we claim that the two genres make use of a single verbal system,
but differ in their preferences for certain parts of that system.
More specifically, we assume that all types of clauses and clause patterns known in Biblical Hebrew
can be used (with the same functionalities) in both prose and poetry,
but that different clause types and patterns are dominant in the two genres.

LAF-Fabric offers an excellent opportunity to test these assumptions
as it provides direct access to the ETCBC database
from which the clause patterns attested in the Hebrew texts can be easily extracted,
as it contains syntactic hierarchies for each chapter of the Hebrew Bible.

As an initial experiment we have created an {\em IPython Notebook} in which we have sorted and counted all
{\em asyndetic} sequences of a mother clause and a daughter clause attested in selected prosaic,
poetic and prophetic sections of the Hebrew Bible.
LAF-Fabric enables us to iterate over all clauses in the preselected texts.
For each clause, we have retrieved the value of its {\em clause atom relation}\footnote{See the comprehensive documentation
of the ETCBC features at \url{http://shebanq-doc.readthedocs.org/en/latest/features/index/code.html}.}
feature,
which is coded as a three-digit value identifying the type of relation between that clause and its mother
(i.e.: asyndetic, parallel, syndetic, coordinate, subordinate, etc.),
the tense of the verbal predicate of the clause (i.e.: imperfect, perfect, imperative, etc.),
and the tense of the verbal predicate of the mother.
Further details on the use of specific functionalities provided by LAF-Fabric and the Python code
written for this task can be found in the notebook
entitled {\em AsyndeticClauseFunctions} by \citeasnoun{Kalkman2013}. 

In table~\ref{freqacpatterns},
we present a top-10 of most frequently attested asyndetic clause patterns in the current ETCBC data
analysed by our LAF-Fabric task.

\begin{table}[htb]
	\centering
	\begin{tabular}{c|rr|rr|rr|rr}
	\toprule
{\small\bfseries CARnumber}&\parbox[t]{8mm}{\small\bfseries Prose}&\parbox[t]{9mm}{\hfill\small\bfseries \%}&\parbox[t]{8mm}{\small\bfseries Poetry}&\parbox[t]{9mm}{\hfill\small\bfseries \%}&\parbox[t]{8mm}{\small\bfseries Prophecy}&\parbox[t]{9mm}{\hfill\small\bfseries \%}&\parbox[t]{8mm}{\small\bfseries Totals}&\parbox[t]{9mm}{\hfill\small\bfseries \%}\\ \hline
nominal $\leadsto$ nominal&429&14.64&493&12.07&328&8.01&\bfseries 1250&\bfseries 11.25\\
imperfect $\leadsto$ imperfect&371&12.66&544&13.32&332&8.11&\bfseries 1247&\bfseries 11.22\\
perfect $\leadsto$ perfect&120&4.09&392&9.6&555&13.55&\bfseries 1067&\bfseries 9.60\\
nominal $\leadsto$ imperfect&232&7.92&250&6.12&244&5.96&\bfseries 726&\bfseries 6.53\\
perfect $\leadsto$ nominal&161&5.49&213&5.21&340&8.3&\bfseries 714&\bfseries 6.43\\
imperfect $\leadsto$ nominal&116&3.96&328&8.03&249&6.08&\bfseries 693&\bfseries 6.24\\
perfect $\leadsto$ imperfect&145&4.95&187&4.58&273&6.67&\bfseries 605&\bfseries 5.45\\
nominal $\leadsto$ perfect&145&4.95&204&4.99&242&5.91&\bfseries 591&\bfseries 5.32\\
imperative $\leadsto$ nominal&54&1.84&270&6.61&212&5.18&\bfseries 536&\bfseries 4.82\\
nominal $\leadsto$ imperative&128&4.37&123&3.01&74&1.81&\bfseries 325&\bfseries 2.93\\
\bfseries Totals&\bfseries 1901&\bfseries 64.86&\bfseries 3004&\bfseries 73.54&\bfseries 2849&\bfseries 69.57&\bfseries 7754&\bfseries 69.79\\
	\bottomrule
	\end{tabular}
	\caption{ Frequency table of attestations of asyndetic clause patterns in Biblical Hebrew prose,
    poetry and prophecy. The percentage column is the fraction for the Clause-Atom-Relation-number within the genre.}
	\label{freqacpatterns}
\end{table}

\begin{figure}[htb]
	\begin{center}
	\includegraphics[scale=0.7]{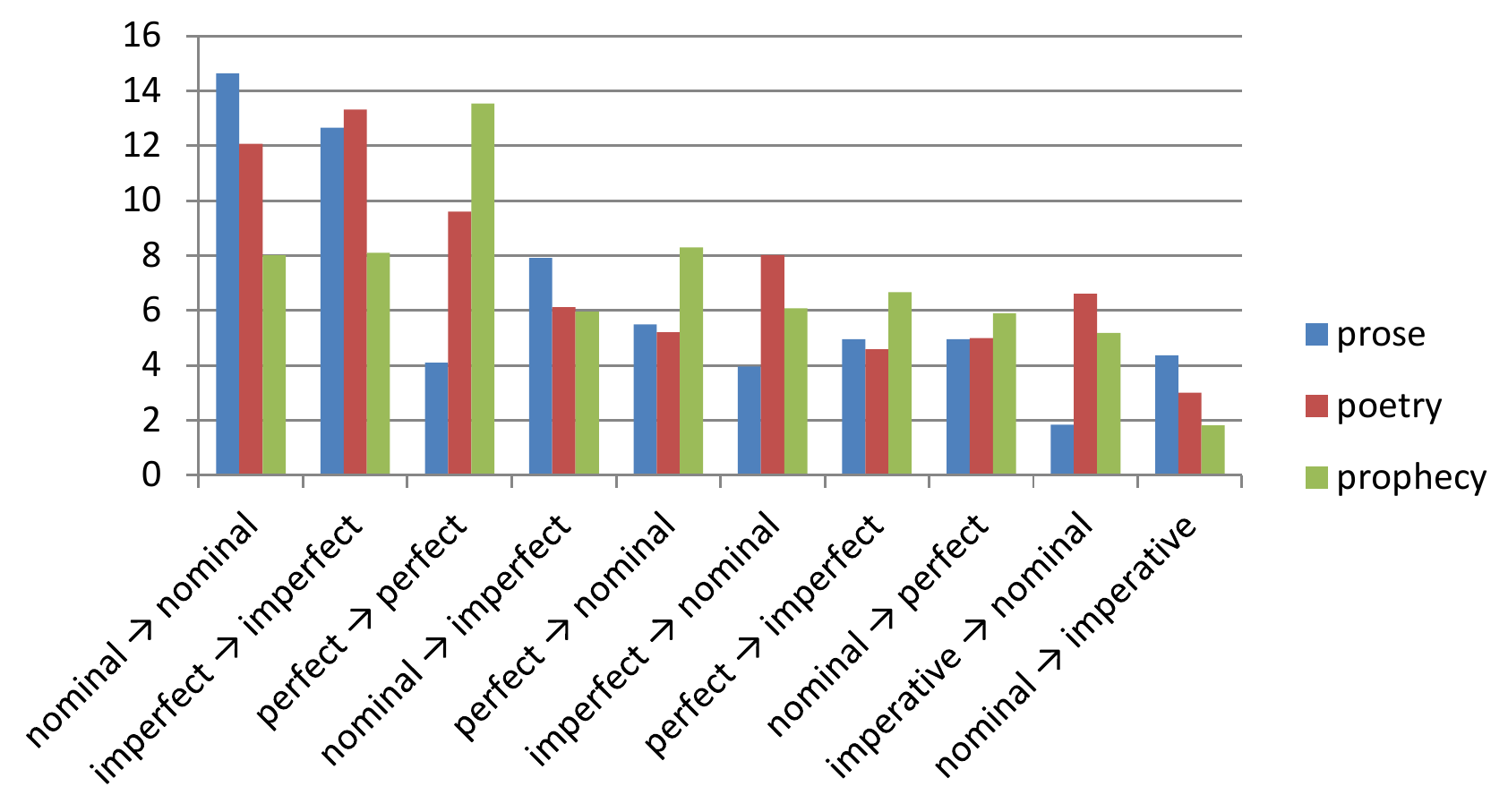}
	\end{center}
	\caption{ Bar graph of frequencies of asyndetic clause patterns in Biblical Hebrew prose, poetry and prophecy.}
	\label{freqacgraph}
\end{figure}

As the table shows, these ten patterns account for over 70\% of all 11,111 patterns that have been found
in the prosaic, poetic and prophetic texts.
Several interesting observations can be made.
First of all, this top-10 does not contain patterns that are strongly attested in one genre,
while being virtually absent in another.
Instead, all of the types of sequences do have quite a number of occurrences in each of the three genres.
(though the rather low number of patterns analysed for poetry forces us to adopt a cautious attitude at this point).

On the other hand, the differences should not be overlooked.
Another visualization of the results may help us in this regard.
In fig~\ref{freqacgraph}, the same statistical data are presented in a bar graph. 
As the graph shows, the pattern {\em perfect} $\leadsto$ {\em perfect}, {\em imperative} $\leadsto$ {\em nominal clause},
and, to a lower extent,
the pattern {\em imperfect} $\leadsto$ {\em nominal clause},
are far better attested in poetry and prophecy than in prose.
Conversely, sequences of two nominal clauses play an important role
in each of the genres, but are most strongly attested in prosaic texts. 
These observations can be explained by referring to the important claim made by several text linguists
studying the Hebrew verbal system that the marking of the mode of communication,
which can be either narrative or discursive (i.e. direct speech) is an important function
of verbal forms in Hebrew \cite[pp.~45-48, 51-52, 55]{Weinrich},
\cite[pp.~182-183]{Schneider}, \cite[pp.~29-34]{Niccacci}.
Narrative discourse is hardly attested in poetry and prophecy,
while it is a dominant mode of communication in prose.
It is therefore not surprising that the patterns {\em perfect} $\leadsto$ {\em perfect},
{\em imperfect} $\leadsto$ {\em nominal clause} and {\em imperative} $\leadsto$ {\em nominal clause},
which are characteristic of direct speech (instead of narrative) discourse,
have higher relative frequencies in poetry and prophecy than in prose.

All in all, we can draw the preliminary conclusion that, while certain patterns are more strongly attested
in one genre than in the others, the differences between the genres in the relative numbers of occurrences
of the asyndetic patterns are not extreme.
This suggests that indeed one linguistic system underlies the functioning of verbal forms and clause types
in the different genres attested in the Hebrew Bible.
Moreover, the main criterion on the basis of which the forms and constructions belonging
to the Hebrew verbal system can be further categorized is not so much that of genre
(prose vs. poetry vs. prophecy),
but rather that of mode of communication (narrative vs. discursive).
To summarize, the use of verbal forms in Hebrew poetry may not at all be as chaotic as grammars,
commentaries and Bible translations seem tot suggest.

This type of experiments conducted with the help of LAF-Fabric has constituted the basis for more
profound research into the Biblical Hebrew verbal system \cite{Kalkman2014,Kalkman2015}.
As part of our research project, we have developed a Java program
in which use is made of the current (May 2014) version of the data included in the ETCBC database.
By concentrating on the syntactic patterns that are identified in the actual data,
the program calculates the default and inherited functions that are to be assigned
to each verbal form in the book of Psalms.
Based on these calculations, the program also offers a translation of the verbal forms
(and other basic constituents, such as subjects and objects) attested in each of the 150 Psalms.
The translations and the results of the calculations made by our program are presented
on a website \cite{Kalkman2015}.
This website also provides a description of our methodology and theories.

All in all, LAF-Fabric proves itself to be an indispensable tool for obtaining new insights
in the grammar of the Biblical Hebrew verb,
as it enables us to systematically extract and collect for further analysis the linguistic patterns
that have a decisive impact on the functioning of verbal forms in Biblical Hebrew.

\section{Data Oriented Parsing of classical Hebrew: generating trees}
There are many unresolved questions concerning the history of composition
and transmission of the Hebrew Bible. One line of research in this area
is to compare and cluster the texts with a view to classifying them
along the dimensions of historical time, geography, and religious
context. Often these classifications are based on intuition and implicit
characteristics. The comparison of syntactic trees could provide this
method with a more objective underpinning and deliver stronger results.

For a start, the LAF data has been exported to syntactic trees, in effect turning the
Hebrew Bible into a treebank for natural language processing (NLP).\footnote{%
	For details on the conversion of the ETCBC data to trees, refer to
	the following IPython notebook:
	\url{http://nbviewer.ipython.org/github/ETCBC/laf-fabric-nbs/blob/master/trees_bhs.ipynb}}
This has led to two applications:

\begin{enumerate*}
	\item extraction of recurring syntactic patterns (tree fragments)
	\item construction and evaluation of a Data-Oriented Parsing grammar from
		said patterns.
\end{enumerate*}

\begin{figure}[htb]
	\begin{center}
	\includegraphics{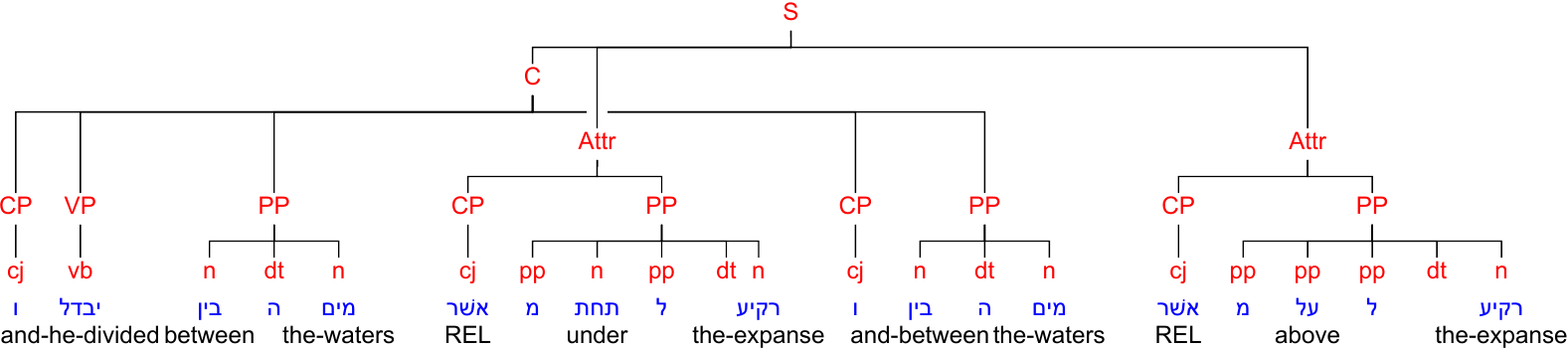}
	\end{center}
	\caption{A sentence with a discontinuous constituent. Genesis chapter 1 verse 7.}
	\label{disctree}
\end{figure}

Since classical Hebrew has a relatively free word order, we make use of
discontinuous constituents in the syntactic trees, inspired by the Negra corpus
annotation~\cite{negra}. See figure~\ref{disctree} for an example of
a sentence with such a discontinuous constituent.
Table~\ref{legend} describes the syntactic categories and Part-of-Speech
tags that appear in the trees.

\begin{table}[htb]
\begin{center}
\begin{tabular}{lllll}
\multicolumn{2}{c}{Syntactic categories} & \hspace{2em} & \multicolumn{2}{c}{Part-of-Speech tags} \\
\cmidrule(r){1-2} \cmidrule(r){4-5}
S     & sentence               & & cj  & conjunction \\
C     & clause                 & & vb  & verb \\
CP    & conjunctive phrase     & & aj  & adjective \\
VP    & verbal phrase          & & n   & noun \\
SU    & subphrase              & & pp  & personal pronoun \\
NP    & nominal phrase         & & dt  & determiner \\
PrNP  & proper noun phrase     & &     & \\
PP    & prepositional phrase   & &     & \\
Attr  & attributive clause     & &     & \\
\end{tabular}
\end{center}
\caption{The syntactic categories and Part-of-Speech tags used in the Hebrew Bible trees.}
\label{legend}
\end{table}

\begin{figure}[htb]
	\begin{center}
	\includegraphics{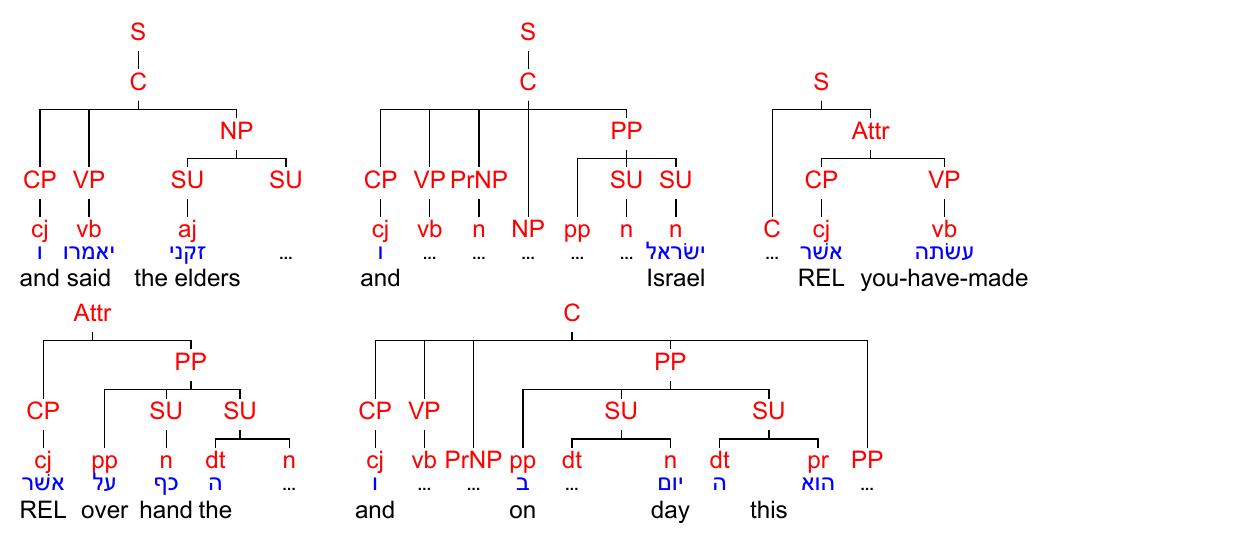}
	\end{center}
	\caption{Some example fragments extracted from
	the Old Testament annotations.}
	\label{figfragments}
\end{figure}

It is possible to extract recurring patterns from a collection of tree structures
using an algorithm first described in \citeasnoun{sangatiefficiently};
we use a faster method that is also able to handle discontinuous
constituents \cite{vancranenburgh2014linear}.
The algorithm compares each pair of tree structures and extracts the largest
fragments they have in common, along with their occurrence counts.
Tree fragments may consist of phrases with or without words,
and the words do not have to form a contiguous phrase.
Since the fragments are found in pairs of trees, this count will always be at
least two. In this way idioms and linguistic constructions can be detected in
syntactic corpora. Figure~\ref{figfragments} shows a sample of fragments
extracted in this way.

\begin{table}[htb]
	\centering
	\begin{tabular}{lr}
	\toprule

	number of sentences:      &   2,000 \\
	longest sentence:         &     19 \\
	labelled f-measure:       &  90.0 \\
	exact match:              &  75.3 \\
	\bottomrule
	\end{tabular}
	\caption{DOP parsing results with the Hebrew Bible.}
	\label{parseresults}
\end{table}

Aside from corpus analysis and stylometry, the extracted fragments can also be
used as a grammar that assigns a syntactic analysis to a given sentence
(parsing). Since classical Hebrew is a dead language, this may at first sight
appear to be a pointless exercise. However, training a probabilistic grammar
and evaluating it gives an impression of how well statistical patterns
in a corpus can be exploited to extrapolate the syntactic structure
of new sentences.

The use of tree fragments as grammar was first proposed in the
Data-Oriented Parsing (DOP) framework \cite{scha1990,bod1992computational}.
We evaluate the performance of parsing classical Hebrew by taking the first
50,000 sentences as training data, and evaluate the resulting grammar on
a held-out set containing the next 2,000 sentences.
In this experiment, the parser is supplied with both words and part-of-speech
tags.
We use the implementation presented in \citeasnoun{Cranenburgh2013}, which
supports trees with discontinuous constituents. For the results, see
Table~\ref{parseresults}.
In the evaluation, the f-measure is the harmonic mean of the precision and recall of
correctly identified constituents~\cite{collins1997three}; the exact match is the
percentage of trees where all constituents are correct.

The results are encouraging, although it should be noted that the
sentences in this held-out set are short.


\section*{Acknowledgements}
The authors are indebted to Wido van Peursen and Rens Bod for setting the scene
for the meeting between theology and computational linguistics, and to
Constantijn Sikkel for enlightening conversations about the ETCBC data.

\bibliographystyle{clin} 
\bibliography{bibliography}  

\end{document}